\documentclass[10pt,twocolumn,letterpaper]{article}

\usepackage{cvpr}              

\usepackage[table]{xcolor}
\usepackage{multirow}
\usepackage{xspace}
\usepackage{amsthm}
\usepackage{float}
\DeclareMathAlphabet\mathbfcal{OMS}{cmsy}{b}{n}

\newcommand{\mat}[1]{\mathbf{#1}}

\theoremstyle{plain}
\newtheorem{theorem}{Theorem}
\theoremstyle{definition}
\newtheorem{definition}[theorem]{Definition}

\definecolor{lightgray}{gray}{0.9}

\definecolor{cvprblue}{rgb}{0.21,0.49,0.74}
\usepackage[pagebackref,breaklinks,colorlinks,allcolors=cvprblue]{hyperref}

\def\paperID{6089} 
\def\confName{CVPR}
\def\confYear{2026}

\title{VEGAS: Mitigating Hallucinations in Large Vision-Language Models via Vision-Encoder Attention Guided Adaptive Steering}

\author{Zihu Wang\\
University of California, Santa Barbara\\
{\tt\small zihu\_wang@ucsb.edu}
\and
Boxun Xu\\
University of California, Santa Barbara\\
{\tt\small boxunxu@ucsb.edu}
\and
Yuxuan Xia\\
University of California, Santa Barbara\\
{\tt\small yuxuanxia@ucsb.edu}
\and
Peng Li\\
University of California, Santa Barbara\\
{\tt\small lip@ucsb.edu}
}

\begin{document}
\maketitle
\begin{abstract}


Large vision–language models (LVLMs) exhibit impressive ability to jointly reason over visual and textual inputs. However, they often produce outputs that are linguistically fluent but factually inconsistent with the visual evidence, i.e., they hallucinate. Despite growing efforts to mitigate such hallucinations, a key question remains: what form of visual attention can effectively suppress hallucinations during decoding? In this work, we provide a simple answer: the vision encoder’s own attention map. We show that LVLMs tend to hallucinate when their final visual-attention maps fail to concentrate on key image objects, whereas the vision encoder’s more concentrated attention maps substantially reduce hallucinations. To further investigate the cause, we analyze vision–text conflicts during decoding and find that these conflicts peak in the language model’s middle layers. Injecting the vision encoder’s attention maps into these layers effectively suppresses hallucinations. Building on these insights, we introduce VEGAS, a simple yet effective inference-time method that integrates the vision encoder’s attention maps into the language model’s mid-layers and adaptively steers tokens which fail to concentrate on key image objects. Extensive experiments across multiple benchmarks demonstrate that VEGAS consistently achieves state-of-the-art performance in reducing hallucinations.

\end{abstract}    
\section{Introduction}
\label{sec:intro}

\begin{figure}[t]
  \centering
   \includegraphics[width=\linewidth]{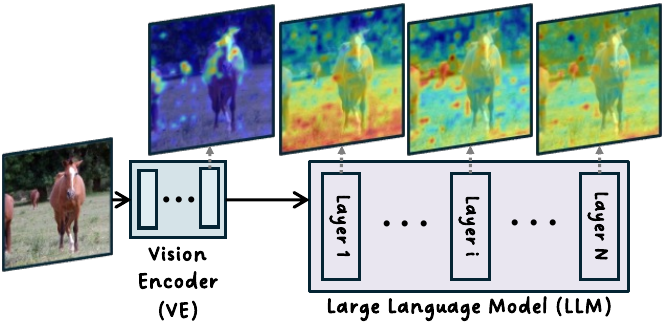}
   \caption{Visualization of visual attention across layers in LLaVA-1.5. The vision encoder’s [CLS] token attention at the model’s final layer shows much tighter focus on major image objects, compared to generated tokens' visual attention in the LLM.}
   \label{fig:vit_vs_llm}
\end{figure}

Driven by recent advances in vision and language modeling, Large Vision–Language Models (LVLMs)~\cite{chen2023shikra, liu2024improved, yin2024survey, zhu2023minigpt4} empower multimodal reasoning by jointly processing images and text. These models are already widely used across applications such as image captioning, conversational assistants, and autonomous systems~\cite{wu2023multimodal,cui2024survey}. Despite their impressive performance, LVLMs often generate responses that are syntactically fluent but visually inconsistent, which is a phenomenon known as hallucination that undermines their reliability in real-world deployments~\cite{liu2024survey}.


Recently, several studies have identified a major source of hallucination in large vision–language models (LVLMs): a vision–language conflict, wherein the model attends disproportionately to textual tokens while under-utilizing visual evidence~\cite{liu2024paying,jiang2025devils,li2025hidden,liu2024survey}. To address this, researchers have proposed methods such as vision attention enhancement~\cite{liu2024paying,jiang2025devils}, latent steering~\cite{li2025hidden}, and contrastive decoding~\cite{wan2025only}. However, a critical question remains under-explored: during the LVLM decoding process, what type of attention distribution over visual tokens minimizes hallucinations?

This paper presents a straightforward answer to the question above: namely, the vision attention distribution produced by the vision encoder (VE). Recent studies~\cite{fu2024hidden,liu2025visual} demonstrate that the VE of an LVLM consistently outperforms the entire model on numerous visual tasks. As illustrated in \cref{fig:vit_vs_llm}, \textbf{the vision encoder’s attention map clearly concentrates on the image’s key objects, whereas the LLM’s visual attention tends to be diffuse and distracted by background details}. To quantify how well an attention map concentrates on salient objects, we introduce the metric Block Entropy (BE). A higher BE indicates poorer concentration, i.e., the attention is spread more uniformly across the image. Using this metric, we show that tokens with higher vision attention BE in the LLM are more likely to be hallucinated. In addition, we observe that the VE’s attention maps consistently exhibit lower BE, reflecting stronger focus on key objects. These findings imply that substituting the LLM’s generated tokens' visual attention with the VE’s attention maps can effectively reduce hallucinations.

Given the aforementioned insights, a natural question follows: at which layers should we integrate the VE attention? To answer this question, we investigate the evolution of text and vision attention in the LLM. By analyzing Vision Attention Ratio (VAR)~\cite{jiang2025devils} and Text-to-visual Entropy Ratio (TVER)~\cite{wan2025only} across layers, it confirms that, \textbf{in the middle layers of the LLM, a model pays highest attention to the image. However, middle layers lack effective visual information.} We thus draw a conclusion: middle layers are a source of LVLM hallucinations. Our studies show that integrating the VE's attention into middle layers can effectively reduce LVLM hallucinations.

With these insights, we propose \textbf{V}ision-\textbf{E}ncoder Attention \textbf{G}uided \textbf{A}daptive \textbf{S}teering (\textbf{VEGAS}), a training-free, inference-time method for mitigating hallucinations in LVLMs. VEGAS integrates the VE’s vision attention maps into the middle layers of the LLM, enabling the model to extract more meaningful and critical visual information. To prevent overemphasis on those major objects in images and neglect of background context, VEGAS introduces an Adaptive Logits Steering mechanism, which combines the original logits with the attention-replaced logits. Specifically, when a newly generated token’s vision attention block entropy (VABE) is high, VEGAS assigns greater weight to the attention-replaced logits. Extensive experiments across multiple benchmarks and LVLM architectures demonstrate that VEGAS achieves state-of-the-art performance in reducing hallucinations.


\section{Related Work}
\label{sec:related_work}

\subsection{Large Vision-Language Models}
\label{ssec:vlms}

The rapid advancement of large language models (LLMs)~\cite{vaswani2017attention,brown2020language,touvron2023llama,chiang2023vicuna} has catalyzed the rise of large vision–language models (LVLMs). Early vision–language models such as VisualBERT~\cite{li2019visualbert} and BLIP~\cite{li2022blip} enabled LLMs to integrate visual information and perform vision–language tasks. More recently, by connecting a vision encoder and an LLM via a connector such as a linear projection~\cite{liu2023visual} or a Q-Former~\cite{li2023blip2}, LVLMs have achieved enhanced reasoning ability—benefiting from visual-instruction tuning techniques~\cite{liu2023visual,liu2024improved}. Despite the strong performance of models like Shikra~\cite{chen2023shikra}, LLaVA~\cite{liu2023visual,liu2024improved}, and MiniGPT-4~\cite{zhu2023minigpt4}, these models still generate outputs that are visually inconsistent with their input images which is a phenomenon known as hallucinations~\cite{liu2024survey}.

\subsection{Mitigation of LVLM Hallucinations}
\label{ssec:hallucination_vlms}

Many recent studies have investigated the causes of hallucinations in LVLMs and proposed corresponding mitigation strategies. A primary source of hallucination stems from biases and noise present in training data~\cite{yu2024hallucidoctor}. To address this issue, researchers have developed methods for high-quality data selection and annotation~\cite{hu2023ciem,liu2023mitigating,gunjal2024detecting,van2022grit}. Beyond improving data quality, modality-matching techniques~\cite{kim2023exposing} and post-training alignment methods~\cite{chen2024dress,sun2024aligning} have been widely adopted to enhance LVLM reliability. While effective, these approaches often require substantial human annotation effort or impose significant computational costs. Another line of work addresses hallucinations by training auxiliary models~\cite{zhou2023analyzing} or leveraging external expert vision and language models~\cite{yin2024woodpecker, wang2023amber, liu2023mitigating} for hallucination detection and correction. However, these methods face practical deployment challenges due to the additional data and computational resources required for training or invoking external models.

To reduce annotation efforts and computational cost in large-scale training, many training-free methods have been proposed. For example, methods such as VCD~\cite{leng2024mitigating}, PAI~\cite{liu2024paying}, ICD~\cite{liu2024reducing}, and DAMRO~\cite{gong2024damro} leverage contrastive decoding between the original logits and logits derived from noisy inputs. In contrast, \cite{cho2025you} ensembles logits produced from different image crops to extract and combine local visual information. Additionally, some other approaches~\cite{liu2024reducing,li2025hidden} steer the latent embeddings towards positive samples to reduce hallucinations.

More recently, studies focus on latent embedding dynamics: for example, PAI~\cite{liu2024paying} method shows that a scale disparity between the vision encoder and LLM drives hallucination and recommends amplifying vision-token attention in decoding; some works observe that the final LLM layer produces more misinformation than intermediate layers~\cite{wang2024mllm,li2025hidden,wang2025damo}; and others reveal that certain attention heads are prone to hallucinations~\cite{jiang2025devils, sarkar2025mitigating}. However, these methods stop short of analyzing hallucination risk in terms of image token attention distributions. In this paper, we dive deep into the visual attention distributions of LVLMs and propose a training-free method to mitigate hallucination without relying on an external expert model.

\section{Preliminary}

\subsection{LVLM decoding}
LVLMs are designed to jointly understand visual and textual inputs. Typically, an LVLM comprises three main components: a vision encoder (VE), a connector, and a large language model (LLM). In the generation process, the VE and connector first convert an image input $\mat{x}_v$ into a sequence of visual tokens. These tokens are then concatenated with the tokenized textual prompt $\mat{x}_t$ and provided as input to the LLM. The probability of generating token $\mat{y}_k$ is given by:
{\fontsize{9pt}{10pt}\selectfont
\begin{equation}
    \mat{y}_k\sim p(\mat{y}_k|\mat{x}_v,\mat{x}_t,\mat{y}_{<k})=\mathrm{softmax}(\mathrm{logits}(\mat{y}_k|\mat{x}_v,\mat{x}_t,\mat{y}_{<k}))
\end{equation}}
where $\mat{y}_{<k}$ denotes the sequence of previously generated tokens and $\mathrm{logits}$ denotes the logits output of the LVLM over the vocabulary.

\subsection{Attention Mechanism in Large Language Models}

In an LVLM, the language decoding is performed by an LLM, which computes attention over its input sequence. Concretely, for attention head $h$ at layer $l$, when generating a token, the attention weights matrix $\mat{A}^{(l,h)}$ for an input sequence of length $n$ is given by:
\begin{equation}
\mat{A}^{(l,h)} = \mathrm{softmax}\!\left(\frac{\mat{Q}^{(l,h)}\,\mat{K}^{(l,h)\top}}{\sqrt{d_{k}}}\right),
\end{equation}
where \(\mat{K}^{(l,h)} \in \mathbb{R}^{n \times d_{k}}, \;\mat{Q}^{(l,h)} \in \mathbb{R}^{1 \times d_{k}},\) and \(\mat{V}^{(l,h)} \in \mathbb{R}^{n \times d_{k}}\) denote the key, query and value matrices for that head, and \(d_{k}\) is the hidden dimension. The head’s output is then $\mat{O}^{(l,h)} = \mat{A}^{(l,h)} \,\mat{V}^{(l,h)}$, i.e., each row of \(\mat{V}^{(l,h)}\) is weighted by the corresponding attention weights in \(\mat{A}^{(l,h)}\).

\section{Method}

\subsection{Visual Concentration is a Key in LVLMs}
\label{sec:vision concentration}


\begin{figure}[h]
  \centering
   \includegraphics[width=0.9\linewidth]{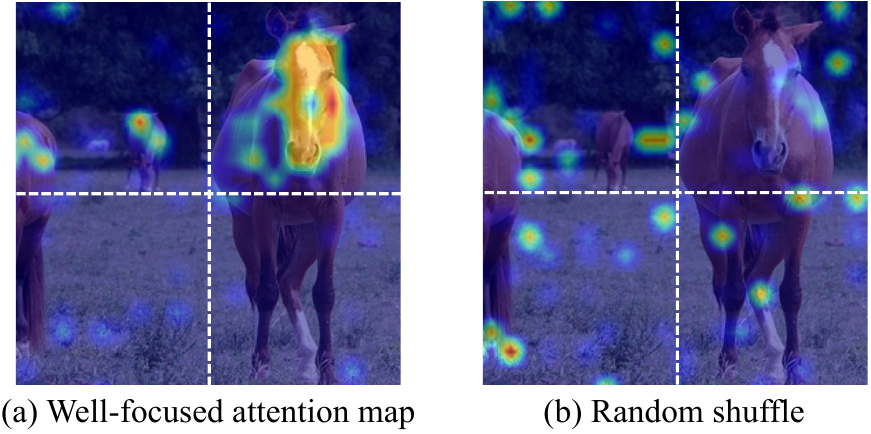}
   \caption{Two attention maps with identical entropy: (a) a well-focused map and (b) its random shuffle. When high-attention values are tightly clustered in one or a few blocks—as in (a)—the block entropy becomes lower, despite the overall entropy remaining the same.}
   \label{fig:ent_vs_be}
\end{figure}

Despite the impressive reasoning capabilities of recent LVLMs, they still often generate outputs that are fluent yet inconsistent with the image contents. Recent studies~\cite{fu2024hidden,liu2025visual} indicate that the vision encoder (VE) of an LVLM often outperforms the full model on many visual tasks. Motivated by this insight, we examine the VE and LLM attention maps over image in \cref{fig:vit_vs_llm}: during decoding, the language model systematically fails to concentrate on the image’s key contents in comparison to the VE. To quantify this discrepancy, we introduce a straightforward and effective metric for attention maps' concentration.

\begin{definition}[Block Entropy]
Given a square matrix $\mat{A} \in \mathbb{R}^{M \times M}$ and a block size $m$ such that $m \mid M$, we partition $\mat{A}$ into $\left(\tfrac{M}{m}\right)^2$ non-overlapping $m \times m$ blocks. Let $\mat{A}_{\text{sum}} \in \mathbb{R}^{(\frac{M}{m})^2}$ denote the vector of blockwise sums, where each entry is the sum of all elements within a block. We normalize $\mat{A}_{\text{sum}}$ using the softmax function:
\begin{equation}
\mat{A}^m = \mathrm{softmax}(\mat{A}_{\text{sum}}) = [A^m_1, A^m_2, \dots, A^m_{(\frac{M}{m})^2}],
\end{equation}
and define the \emph{block entropy} of $\mat{A}$ at block size $m$ as
\begin{equation}
\mathrm{BE}_m(\mat{A}) = -\sum_{i=1}^{(\frac{M}{m})^2} A^m_i \log A^m_i.
\end{equation}
\end{definition}

\begin{figure}[h]
  \centering
   \includegraphics[width=0.9\linewidth]{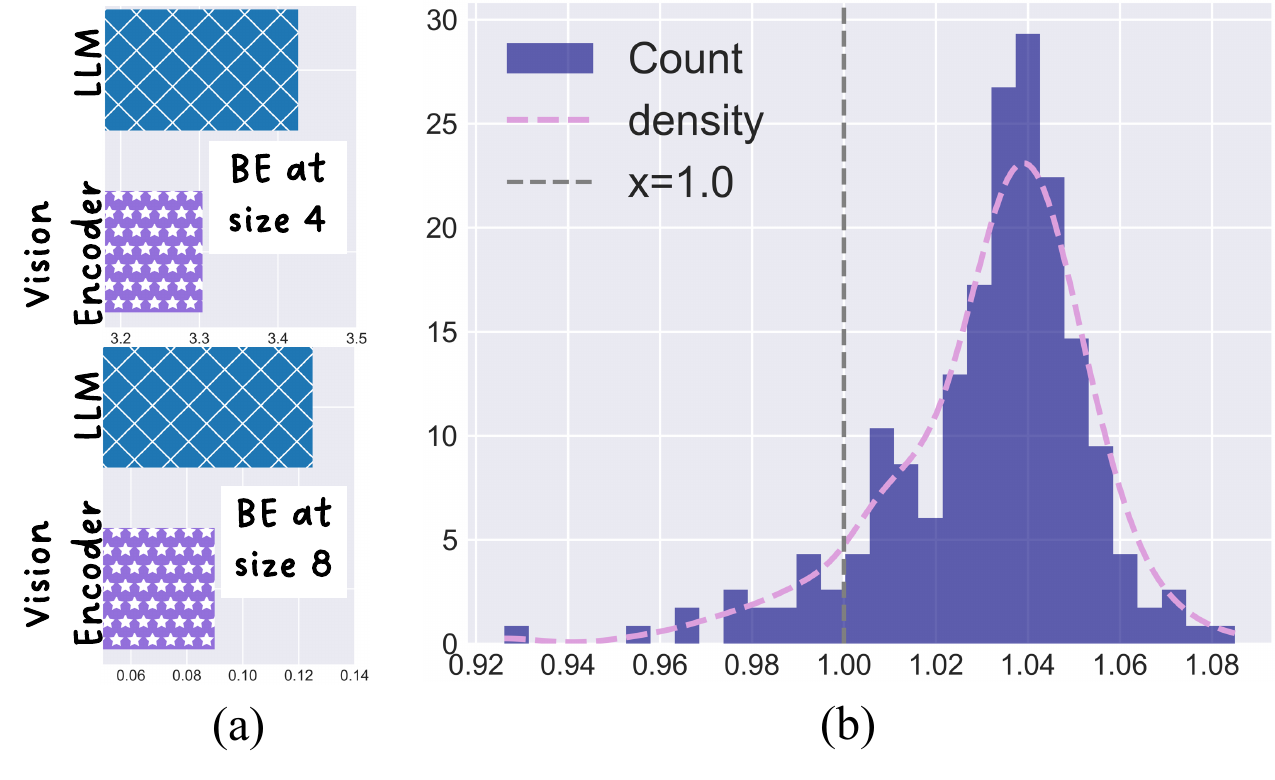}
   \caption{All statistics are averaged over real object tokens in LLaVA-1.5. (a) Comparison of vision attention block entropies, $\mathrm{BE}_4$ and $\mathrm{BE}_8$, of the LLM’s last layer an the VE's last layer. Tokens in the LLM typically exhibit higher block entropy than those from the VE. (b) Ratio of hallucinated-token to non-hallucinated-token block entropy ($\mathrm{BE}_4$) for vision attention maps at the LLM's layer 15. The ratio typically exceeds 1, indicating that hallucinated tokens tend to exhibit larger block entropy. These patterns remain consistent across multiple LLM layers.}
   \label{fig:be_stats}
\end{figure}

Compared with standard entropy, block entropy accounts for the clustering of high-attention values by summing over image blocks. As illustrated in \cref{fig:ent_vs_be}, even though two attention distributions may share the same entropy, the one with high values more tightly clustered around the major object yields a lower block entropy. To further highlight the advantage of this metric, \cref{fig:be_stats}(a) compares the block-entropy values of vision attention maps from the VE and the LLM. The VE’s stronger focus on the key objects in an image leads to consistently lower block entropy than that of the LLM.

Knowing that in LVLMs the VE typically exhibits stronger visual concentration than the LLM, an intriguing question arises: Does the LLM's low visual concentration hint at hallucinations in LVLMs? As shown in \cref{fig:be_stats}(b), on average hallucinated tokens display higher block entropy in their corresponding vision attention maps compared to non-hallucinated tokens. Thus, high block entropy in the LLM vision attention generally serves as a red flag for hallucinations in LVLM outputs. A natural implication follows: one can leverage the high-concentration (low BE) attention maps of the VE to guide decoding in the LLM and thereby reduce hallucinations.




\begin{figure}[h]
  \centering
   \includegraphics[width=\linewidth]{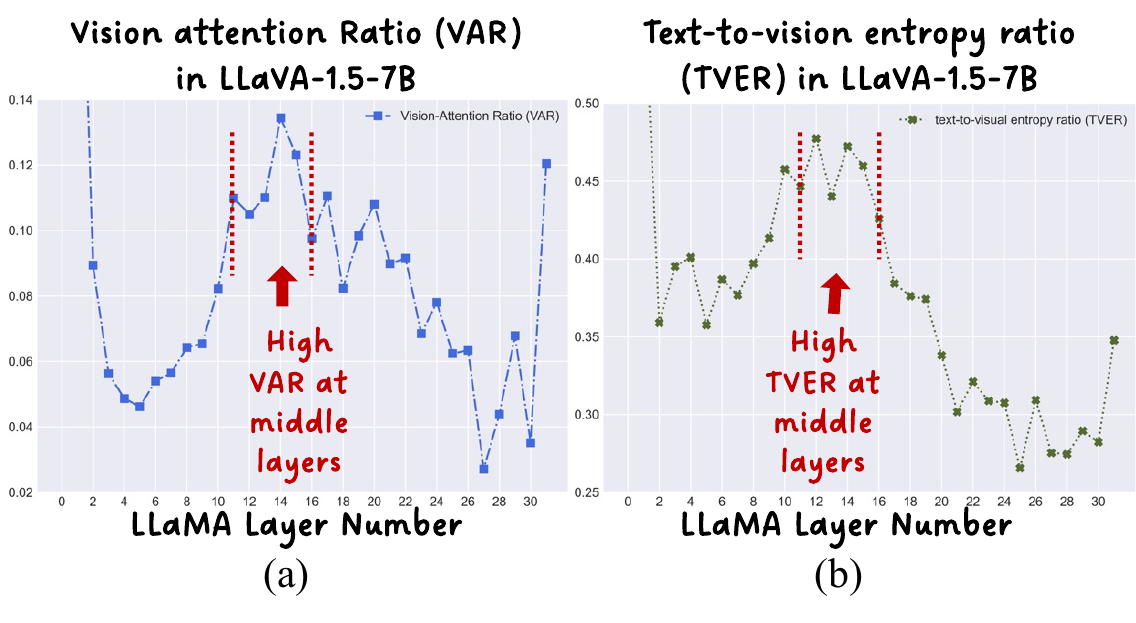}
   \caption{The (a) VAR and (b) TVER across layers in LLAVA‑1.5, averaged over real object tokens. Although visual attention (VAR) peaks in the middle layers, the correspondingly high TVER indicates these layers fail to extract effective visual information.}
   \label{fig:var_tver}
\end{figure}

\subsection{High Visual Attention, Minimal Visual Information in Middle Layers}

To leverage the high-concentration attention maps from the VE during LLM decoding, a straightforward approach is to replace the LLM’s final-layer vision attention with that of the VE. Although our empirical results confirm that this substitution can reduce hallucinations, it may potentially disrupt the image–text alignment already established in earlier layers. To better understand at which layers visual information from the VE is most beneficial, we analyze the evolution of image and text attention patterns across LLM layers.

\paragraph{Middle layers focus most on image tokens.}
We first quantify how much each LLM layer attends to image tokens using the vision-attention ratio (VAR)~\cite{jiang2025devils}:
\begin{equation}
\mathrm{VAR}^l(\mat{y}_k)=\frac{1}{H}\sum_h\sum_{i=1}^{N_v}a^{(l,h)}_k(\mat{v}_i),
\end{equation}
where $a^{(l,h)}_k(\mat{v}_i)$ denotes the normalized attention weight from the $k$-th generated token $\mat{y}_k$ to the $i$-th image token $\mat{v}_i$ at layer $l$ and head $h$. Summing over all image tokens and then averaging over all $H$ heads measures how strongly layer $l$ attends to visual information. As shown in \cref{fig:var_tver}(a), the middle layers exhibit the highest VAR, indicating that they place the greatest emphasis on visual tokens.

\paragraph{Middle layers lack effective visual information.}
Next, we examine the text-to-visual entropy ratio (TVER)~\cite{wan2025only}, which captures the degree of modality bias:
\begin{equation}
\mathrm{TVER}^l(\mat{y}_k)=\sum_h
\frac{\sum_i p^{txt}_{(l,h,k)}(\mat{t}_i)\log p^{txt}_{(l,h,k)}(\mat{t}_i)}
{\sum_i p^{img}_{(l,h,k)}(\mat{v}_i)\log p^{img}_{(l,h,k)}(\mat{v}_i)},
\end{equation}
where $p^{txt}_{(l,h,k)}$ and $p^{img}_{(l,h,k)}$ are the normalized attention distributions of newly generated token $\mat{y}_k$ over text tokens $\mat{t}_i$ and image tokens $\mat{v}_i$, respectively, at layer $l$ and head $h$. A higher TVER indicates a stronger text bias~\cite{wan2025only}, meaning the model extracts relatively less information from the image. As shown in \cref{fig:var_tver}(b), middle layers exhibit the highest TVER, suggesting that they rely more on text and fail to incorporate sufficient visual information.

\begin{figure}[h]
  \centering
   \includegraphics[width=\linewidth]{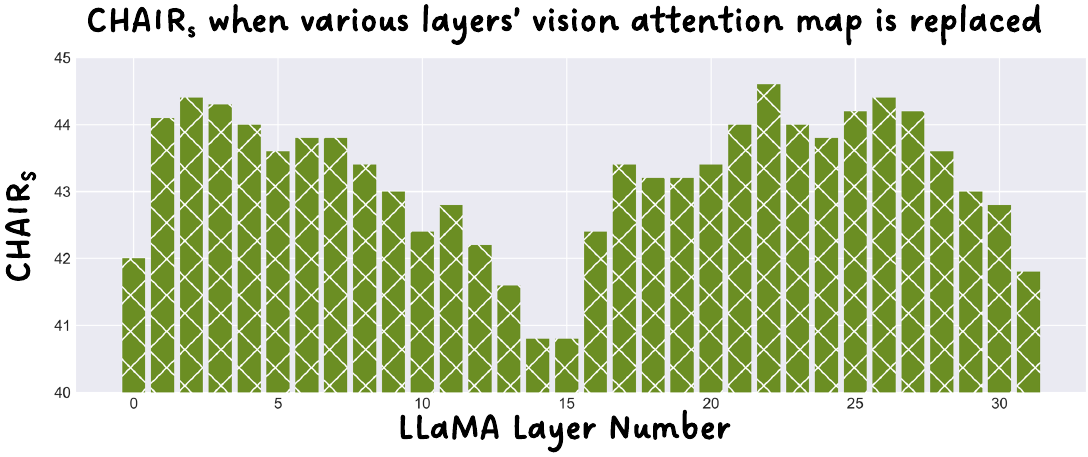}
   \caption{$\mathrm{CHAIR}_S$~\cite{rohrbach2018object} values when the vision attention maps at different layers of the LLM are replaced with the VE’s attention map. The results show that injecting the VE’s attention into the middle layers yields the greatest reduction in hallucinations.}
   \label{fig:layers_replaced}
\end{figure}

\begin{figure*}[ht]
  \centering
   \includegraphics[width=0.95\linewidth]{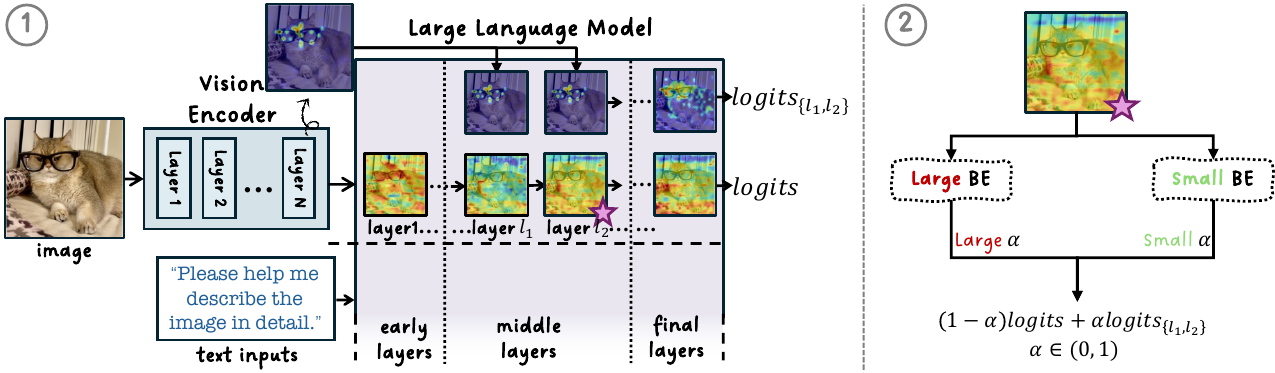}
   \caption{Overview of VEGAS. VEGAS integrates the vision encoder’s attention maps into the LLM’s middle layers and employs adaptive logits steering to reduce hallucinations. It combines the original logits with attention replaced logits to maintain both object focus and background context.}
   \label{fig:overview}
\end{figure*}

\paragraph{Injecting vision attention into middle layers reduces hallucinations.}
The above two findings reveal a critical insight: although the middle layers of the LLM assign the greatest attention to image tokens, they fail to encode effective visual information. To validate this, we replace the LLM’s attention over image tokens at different layers with the attention maps of the VE.

For head $h$ in layer $l$, let $\mat{\tilde{A}}^{(l,h)} \in \mathbb{R}^{n}$ denote the pre-softmax attention over the input sequence of length $n$. The segment corresponding to image tokens is extracted as
\begin{equation}
\mat{\tilde{A}}^{(l,h)}_v = \mat{\tilde{A}}^{(l,h)}[i_s : i_e + 1],
\end{equation}
where $i_s$ and $i_e$ are the start and end indices of the image tokens in the input sequence.  
We replace this segment with the VE’s $[\mathrm{CLS}]$ token attention over image tokens at the last layer, denoted by $\mat{\tilde{A}}^{(h)}_{\mathrm{VE}}$. Because prior work~\cite{darcet2023vision} shows that vision transformers produce extremely high attention values on low-informative image tokens, we clamp the highest values in the attention map to the average visual attention value. And to preserve the overall vision-attention ratio (VAR) after substitution, we apply the following adjustment:
\begin{equation}
\mat{\tilde{A}}^{(l,h)}_v \leftarrow 
\mat{\tilde{A}}^{(h)}_{\mathrm{VE}}
- \mathrm{mean}\!\left(\mat{\tilde{A}}^{(h)}_{\mathrm{VE}}\right)
+ \mathrm{mean}\!\left(\mat{\tilde{A}}^{(l,h)}_v\right).
\end{equation}
Since the LLM typically contains more attention heads than the VE, we replicate $\mat{\tilde{A}}^{(h)}_{\mathrm{VE}}$ repeatedly to match the count of LLM heads.

As shown in \cref{fig:layers_replaced}, integrating the VE’s attention into the most image-focused yet visually-deficient middle layers yields the greatest reduction in hallucinations.




\subsection{VEGAS: Vision Encoder Attention Guided Adaptive Steering}

\paragraph{Logits Steering.}
Integrating the VE's attention map into the LLM can effectively mitigate hallucinations. However, while the VE’s attention provides strong focus on major objects, it may cause the LLM to overlook contextual or background details. To balance these effects, we combine the original logits and the attention-replaced logits to achieve optimal performance through the following \emph{logits steering} formulation:
\begin{equation}
    \mathrm{logits}^{\prime} = (1-\alpha)\,\mathrm{logits} + \alpha\,\mathrm{logits}_{\{l_1,l_2,\dots\}},
\end{equation}
where $\mathrm{logits}^{\prime}$ denotes the final logits used for vocabulary sampling, $\mathrm{logits}$ represents the original LLM logits, and $\mathrm{logits}_{\{l_1,l_2,\dots\}}$ corresponds to the logits computed when the visual-attention maps of layers $l \in \{l_1,l_2,\dots\}$ are replaced with those from the VE. The scalar $\alpha \in (0,1)$ controls the balance between the two logits. To amplify the impact of the VE’s attention, we follow~\cite{liu2024paying} to apply a visual attention enhancement to $\mathrm{logits}_{\{l_1,l_2,\dots\}}$.

\paragraph{Adaptive Logits Steering.}
As discussed in \cref{sec:vision concentration}, high block entropy in the LLM’s visual attention often indicates potential hallucinations. We thus use the vision attention block entropy (VABE) as an adaptive indicator:
\begin{equation}
\label{eq:vabe}
\mathrm{VABE}_4^l = \frac{1}{H} \sum_{h=1}^{H} \mathrm{BE}_4\!\left(\mat{\tilde{A}}^{(l,h)}_v\right),
\end{equation}
where $H$ is the number of attention heads and $\mat{\tilde{A}}^{(l,h)}_v$ denotes the pre-softmax attention over image tokens at head $h$ of layer $l$.  

Finally, the weighting coefficient $\alpha$ in the above logits steering is adaptively determined as:
\begin{equation}
\label{eq:adaptive_alpha}
    \alpha =
\begin{cases}
\alpha_{1}, & \text{if } \mathrm{VABE}_4^l > \eta,\\[4pt]
\alpha_{2}, & \text{otherwise},
\end{cases}
\end{equation}
where $\eta$ is a threshold controlling when to apply stronger steering.



\begin{table*}[t]
    \centering
    \caption{CHAIR hallucination evaluation results. Maximum new token is set to 512.}
    \scalebox{0.84}{
    \setlength{\tabcolsep}{4.6mm}{
    \begin{tabular}{llcccccc}
        \toprule
        \multirow{2}{*}{\centering Decoding} & \multirow{2}{*}{\centering Method} & \multicolumn{2}{c}{LLAVA-1.5~\cite{liu2023visual}} & \multicolumn{2}{c}{MiniGPT-4~\cite{zhu2023minigpt4}} & \multicolumn{2}{c}{Shikra~\cite{chen2023shikra}} \\
        \cmidrule(lr){3-4} \cmidrule(lr){5-6} \cmidrule(lr){7-8} 
        & & CHAIR$_\text{S}\downarrow$ & CHAIR$_\text{I}\downarrow$ & CHAIR$_\text{S}\downarrow$ & CHAIR$_\text{I}\downarrow$ & CHAIR$_\text{S}\downarrow$ & CHAIR$_\text{I}\downarrow$ \\
        \midrule
        \multirow{5}{*}{Greedy} 
        & Vanilla & 43.8 & 13.0 & 33.8 & 10.4 & 54.6 & 15.0 \\
        & VCD~\cite{zhang2023mitigating} & 43.5 & 13.8 & - & - & - & -\\
        & PAI~\cite{liu2024paying} & 28.8 & 7.8 & 24.0 & 9.0 & 31.6 & 8.9   \\
        & ~\cite{jiang2025devils} & 29.5 & 8.8 & 23.4 & 8.8 & \textbf{23.4} & \textbf{8.1} \\
        & VISTA~\cite{li2025hidden} & 27.2 & 7.3 & 22.5 & 8.7 & 31.4 & 8.6\\
        \cmidrule(lr){2-8}
        & \textbf{VEGAS} (ours) & \textbf{24.8} & \textbf{7.1} & \textbf{21.4} & \textbf{8.4} & 24.0 & \textbf{8.1}  \\
        \midrule
        \multirow{5}{*}{Beam Search} 
        & Vanilla & 49.8 & 14.0 & 34.6 & 10.1 & 53.4 & 14.1  \\
        & VCD~\cite{zhang2023mitigating} & 50.0 & 14.4 & - & - & - & - \\
        & PAI~\cite{liu2024paying} & 27.5 & 7.8 & 31.8 & 9.9 & 36.2 & 9.8\\
        & ~\cite{jiang2025devils} & 29.4 & 8.5 & 29.8 & 8.8 & \textbf{23.5} & 9.1 \\
        & VISTA~\cite{li2025hidden} & 24.0 & 7.5 & 22.8 & 8.2 & 33.0 & 9.6\\
        \cmidrule(lr){2-8}
        & \textbf{VEGAS} (ours) & \textbf{23.2} & \textbf{7.0} & \textbf{21.7} & \textbf{8.0}  & 24.5 & \textbf{8.8}  \\
        \bottomrule
    \end{tabular}
    }}
    \label{tab:chair}
\end{table*}

\begin{table*}[t]
    \centering
    \caption{Hallucination evaluation results on the POPE benchmark for greedy decoding across three ground-truth label splits.}
    \scalebox{0.84}{
    \setlength{\tabcolsep}{6mm}{
    \begin{tabular}{llcccccc}
        \toprule
        \multirow{2}{*}{\centering Setting} & \multirow{2}{*}{\centering Method} & \multicolumn{2}{c}{LLAVA-1.5~\cite{liu2023visual}} & \multicolumn{2}{c}{MiniGPT-4~\cite{zhu2023minigpt4}} & \multicolumn{2}{c}{Shikra~\cite{chen2023shikra}} \\
        \cmidrule(lr){3-4} \cmidrule(lr){5-6} \cmidrule(lr){7-8} 
        & & Accuracy$\uparrow$ & F1$\uparrow$ & Accuracy$\uparrow$ & F1$\uparrow$ & Accuracy$\uparrow$ & F1$\uparrow$ \\
        \midrule
        \multirow{5}{*}{Random} 
        & Vanilla & 89.33 & 89.29 & 82.33 & 80.64 & 83.36 & 83.52 \\
        & VCD~\cite{zhang2023mitigating} & 89.05 & 89.03 & - &- & - &-\\
        & PAI~\cite{liu2024paying} & 90.03 & 89.98 & 82.30 & 80.73 & 83.30 & 83.53   \\
        & VISTA~\cite{li2025hidden} & 90.03 & \textbf{90.02} & 83.50 & 81.43 & 84.38 & 84.01\\
        \cmidrule(lr){2-8}
        & \textbf{VEGAS} (ours) & \textbf{90.10} & 89.98 & \textbf{84.07} & \textbf{81.58} & \textbf{84.73} & \textbf{84.17}\\
        \midrule
        \multirow{5}{*}{Popular} 
        & Vanilla & 85.90 & 86.32 & 74.93 & 74.66 & 82.67 & 82.89 \\
        & VCD~\cite{zhang2023mitigating} & 85.88 & 86.03 & - &- & - &- \\
        & PAI~\cite{liu2024paying} & 86.06 & 86.42 & 75.80 & 75.40 & 82.55 & 82.80  \\
        & VISTA~\cite{li2025hidden} & 86.73 & 87.15 & 76.48 & 75.03 & 83.27 & 83.34\\
        \cmidrule(lr){2-8}
        & \textbf{VEGAS} (ours) & \textbf{87.30} & \textbf{87.37} & \textbf{77.60} & \textbf{75.93} & \textbf{84.87} & \textbf{84.93} \\
        \midrule
        \multirow{5}{*}{Adversarial} 
        & Vanilla & 80.03 & 81.22 & 71.13 & 71.96 & 78.68 & 79.75 \\
        & VCD~\cite{zhang2023mitigating} & 79.95 & 81.17 & - &- & - &- \\
        & PAI~\cite{liu2024paying} & 81.05 & 82.17 & 71.70 & 72.39 & 78.68 & 79.78 \\
        & VISTA~\cite{li2025hidden} & 81.30 & \textbf{82.70} & 72.47 & 72.78 & 78.63 & 79.00\\
        \cmidrule(lr){2-8}
        & \textbf{VEGAS} (ours) & \textbf{81.43} & 81.78 & \textbf{72.77} & \textbf{73.03} & \textbf{78.94} & \textbf{79.95} \\
        \bottomrule
    \end{tabular}
    }}
    \label{tab:pope}
\end{table*}

\section{Experiments}


In this section, we present experiments across multiple LVLM architectures, various decoding strategies, and diverse benchmarks. Additionally, we conduct comprehensive ablation studies to further demonstrate the effectiveness of VEGAS. In all tables \textbf{bold values} denote the best performance in the corresponding tasks.

\subsection{Experimental Setup}
\paragraph{Models.} We implement and evaluate VEGAS on three representative LVLMs. LLAVA-1.5-7B~\cite{liu2024improved} and Shikra~\cite{chen2023shikra} both adopt a linear projection layer to connect the VE with the LLM. In contrast, MiniGPT-4~\cite{zhu2023minigpt4} incorporates a Q-former~\cite{li2023blip2} to align different modalities.

\paragraph{Implementation Details.}
For LLaVA-1.5 and Shikra, we extract the [CLS]-token's attention over all image tokens from the vision transformer's final layer and inject this attention map into layers 14 and 15 (0-indexed) of the LLM. For MiniGPT-4, we utilize the Q-Former's last cross-attention layer: for each query token, we compute its average attention over all image tokens, then aggregate these averages to form an attention map, which we inject into the LLM's layers 14 and 15. We select $\mathrm{VABE_4^{15}}$ as the hallucination indicator. Additional experimental setup details are provided in the Appendix. All experiments are conducted on a single NVIDIA A100 (80GB) GPU.

\subsection{Main Results}

\paragraph{CHAIR.} 
Caption Hallucination Assessment with Image Relevance (CHAIR)~\cite{rohrbach2018object} is a widely used benchmark for evaluating object hallucinations. A hallucination is defined as a case where the model mentions an object that does not appear in the ground-truth labels. Two metrics are reported: $\mathrm{CHAIR}_S$, the proportion of hallucinated sentences among all generated sentences, and $\mathrm{CHAIR}_I$, the proportion of hallucinated objects among all mentioned objects. Following~\cite{huang2024opera}, we randomly sample 500 images from the MS-COCO 2014 validation set and use the prompt ``\texttt{Please help me describe the image in detail.}''

As shown in \cref{tab:chair}, VEGAS achieves superior performance across all evaluated models in this open-ended image describing task. These results demonstrate that the VE attention map, combined with our proposed adaptive logits steering mechanism, consistently reduces hallucinations across diverse model architectures and decoding strategies.

\paragraph{POPE.} 
Polling-based Object Probing Evaluation (POPE)~\cite{li2023evaluating} assesses hallucinations by prompting the model with the question ``\texttt{Is there a <object> in the image?}'' Here, ``\texttt{<object>}'' is drawn from three label splits: \emph{random}, \emph{popular} (frequently occurring), and \emph{adversarial} (challenging) categories. Following \cite{liu2024paying}, we evaluate on 500 images from the COCO dataset, with six questions per image for each label split.

As demonstrated in \cref{tab:pope}, VEGAS attains superior performance across all evaluated models and label splits. By leveraging the VE attention map alongside adaptive logits steering, the model accurately focuses on queried objects while maintaining awareness of minor background elements when responding to object existence queries.

\begin{figure}[h]
  \centering
   \includegraphics[width=\linewidth]{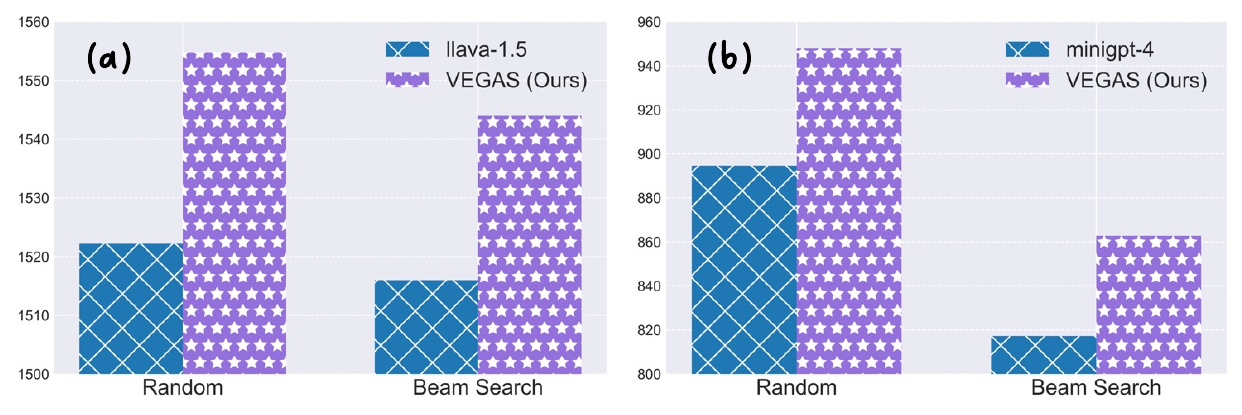}
   \caption{Overall scores on all 14 subtasks of the MME benchmark, comparing with base models using different decoding strategies in (a) LLAVA-1.5 and (b) MINIGPT-4.}
   \label{fig:mme_eval}
\end{figure}

\begin{figure*}[t]
  \centering
   \includegraphics[width=0.95\linewidth]{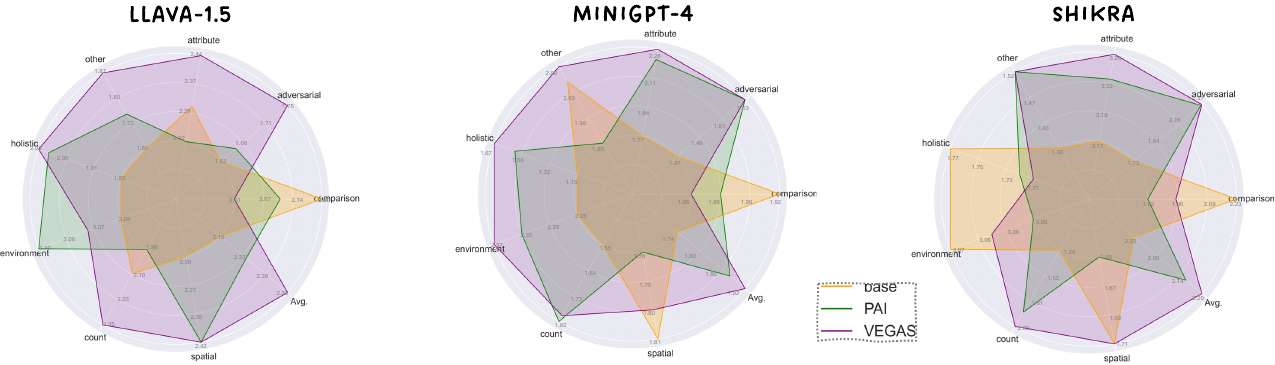}
   \caption{Evaluation results on the eight question categories of MMHal-Bench, where answers are scored by GPT-4. The figure also reports the averages across all categories.}
   \label{fig:mmhal}
\end{figure*}

\paragraph{MME.} MME \cite{fu2025mme} is a benchmark that evaluates a model’s performance across 14 perception and cognition subtasks, providing a comprehensive assessment of multimodal capabilities. We evaluate VEGAS across all 14 subtasks and report overall scores. As shown in \cref{fig:mme_eval}, when compared against the base models using different decoding strategies, VEGAS significantly improves performance by guiding the LLM’s intermediate-layer visual attention to critical contents in images.

\paragraph{MMHal-Bench.} MMHal-Bench~\cite{sun2024aligning} uses 96 The benchmark MMHal‑Bench~\cite{sun2024aligning} comprises 96 image–question pairs covering eight categories—including object attributes, adversarial objects, comparisons, counting, spatial relations, environment, holistic descriptions, and others. Model responses are evaluated by GPT‑4~\cite{achiam2023gpt} to assess hallucination tendencies. In our experiments, we use greedy decoding for all models. \cref{fig:mmhal} demonstrates the performance of base models, PAI~\cite{liu2024paying}, and VEGAS. Across all three base model architectures, VEGAS consistently achieves the best overall performance. Thanks to the vision encoder’s better focus on primary objects and VEGAS’s adaptive logits steering, which preserves critical background details and global image context, our method significantly outperforms existing techniques on this comprehensive VQA benchmarking task.

\subsection{Ablation Study}

\begin{table}[h]
    \centering
    \caption{Ablation study results showing the impact of integrating VE attention into different LLM layers within the VEGAS framework. CHAIR results on LLAVA-1.5 using greedy decoding is reported.}
    \scalebox{0.9}{
    \setlength{\tabcolsep}{2.5mm}{
    \begin{tabular}{lccccc}
        \toprule
        Layers & \{0\} & \{14\} & \{15\} & \{31\} & $\{14,15\}$ \\
        \midrule
        CHAIR$_\text{S}\downarrow$ & 35.5 & 33.0 & 32.2 & 34.8 & \textbf{24.8}\\
        \bottomrule
    \end{tabular}
    }}
    \label{tab:abl_layer}
\end{table}

\paragraph{Layers to integrate VE attention.} 
\cref{fig:layers_replaced} compares the effect of replacing different LLM layers’ attention maps with the vision-encoder (VE) attention. The results reveal that substituting the middle layers, specifically Layer 14 and Layer 15, yields the strongest reduction in hallucinations. Note that those results were obtained without the full VEGAS framework (i.e., without adaptive logits steering). To further study layer selection within the complete VEGAS pipeline, we vary the set of layers ${\{l_1,l_2,\dots\}}$ for which $\mathrm{logits}_{\{l_1,l_2,\dots\}}$ are computed. As shown in \cref{tab:abl_layer}, replacing just Layers 14 and 15 vision attention continues to deliver good performance, and replacing both of them leads to the best results.

\begin{table}[h]
    \centering
    \caption{Impact of the head alignment approaches when integrating VE attention to the LLM. CHAIR results on LLAVA-1.5 using greedy decoding is reported.}
    \scalebox{0.9}{
    \setlength{\tabcolsep}{2.5mm}{
    \begin{tabular}{lccc}
        \toprule
        Head Alignment & \textit{broadcast} & \textit{random} & \textit{similarity} \\
        \midrule
        CHAIR$_\text{S}$/CHAIR$_\text{I}\downarrow$ & 24.8/7.1 & 24.8/7.1 & 25.4/7.8 \\
        \bottomrule
    \end{tabular}
    }}
    \label{tab:abl_head}
\end{table}

\paragraph{Head Alignment.}  
When integrating the vision encoder’s (VE) attention into the LLM, we must align the VE attention maps to match the number of attention heads in the LLM. We evaluate three head-alignment strategies: (1) \textit{broadcast}: replicate the VE attention map repeatedly until its count equals the LLM's head count; (2) \textit{random}: for each LLM head, randomly select and apply a VE head’s attention map; (3) \textit{similarity}: for each LLM head, compute the cosine similarity between that head’s original attention map and each VE head map, then replace with the VE map having the highest similarity. As shown by \cref{tab:abl_head}, the different head alignment approaches lead to close performance. Therefore, for simplicity and efficiency, we adopt the \textit{broadcast} strategy in all other experiments.

\begin{figure}[h]
  \centering
   \includegraphics[width=\linewidth]{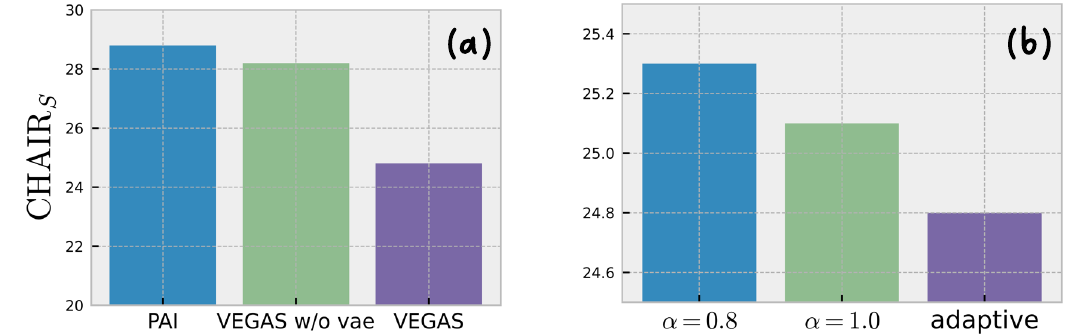}
   \caption{Ablation study results on LLaVA-1.5 with greedy decoding: (a) visual attention enhancement ablation; (b) adaptive logits steering ablation.}
   \label{fig:ablation_study}
\end{figure}

\paragraph{Visual attention enhancement.} Since we adopt the visual‐attention enhancement technique from \cite{liu2024paying} when producing $\mathrm{logits}_{\{l_1,l_2,\dots\}}$, we conduct an ablation study to isolate its impact. In \cref{fig:ablation_study}(a), we compare three configurations: VEGAS, a variant ``VEGAS w/o vae" (where $\mathrm{logits}_{\{l_1,l_2,\dots\}}$ is produced without applying visual attention enhancement), and PAI~\cite{liu2024paying}. The results demonstrate that even without visual attention enhancement, VEGAS w/o vae outperforms PAI in reducing hallucinations. Furthermore, by enhancing the model's attention on the integrated VE attention maps, VEGAS improves focus on critical visual details and achieves additional hallucination reduction.

\paragraph{Adaptive logits steering.}
VEGAS introduces adaptive logits steering to prevent LVLMs from overemphasizing major objects in images. To evaluate this component, we conduct an ablation study comparing different steering strategies. In \cref{fig:ablation_study}(b), we compare VEGAS implementations with fixed logits weight $\alpha$ against our adaptive approach, where $\alpha=1.0$ when VABE is large (indicating higher hallucination risk) and $\alpha=0.8$ when VABE is small. Using the adaptive $\alpha$ achieves the optimal result, confirming that adaptive logits steering is a good strategy to balance the models attention on major objects and background details in an image.

\begin{table}[h]
    \centering
    \caption{Throughput (tokens/second) comparison on LLAVA-1.5-7B using greedy decoding.}
    \scalebox{0.8}{
    \setlength{\tabcolsep}{0.9mm}{
    \begin{tabular}{lccccc}
        \toprule
        Methods & LLAVA-1.5 & VCD~\cite{leng2024mitigating} & OPERA~\cite{huang2024opera} & PAI~\cite{liu2024paying} & VEGAS \\
        \midrule
        Tokens/sec.$\uparrow$ & 34.7 & 18.1 & 12.9 & 26.6 & 25.3 \\
        \bottomrule
    \end{tabular}
    }}
    \label{tab:throughput}
\end{table}

\paragraph{Throughput.} \cref{tab:throughput} compares the inference throughput of VEGAS against several existing state-of-the-art methods. Although VEGAS integrates the VE’s attention maps into the LLM, the original logits and the attention-replaced logits can be computed in parallel, enabling the method to maintain high efficiency.



\section{Discussion}

In this work, we demonstrate that, compared with the large language model (LLM) of a large vision–language model (LVLM), the vision encoder (VE) consistently produces attention maps that are better focused on key objects in the image. Using our proposed metric, Block Entropy (BE), we show that low concentration of an LLM’s visual attention map frequently signals a higher risk of hallucination. By analyzing the evolution of image and text attention across layers, we further observe that the middle layers of the LLM allocate the highest attention to visual tokens, they nevertheless fail to extract meaningful underlying information from the images. Building on these insights, we propose VEGAS: a training-free, inference-time method that integrates the VE’s attention into the LLM’s middle layers and adaptively steers the final logits to reduce hallucination. Extensive experiments across multiple LVLM models, decoding strategies, and benchmarks confirm that VEGAS achieves state-of-the-art performance in mitigating hallucinations. More experiment results and detailed experiment settings are provided in the Appendix.

Despite its effectiveness and efficiency, VEGAS does incur modest additional computational overhead compared to the base foundation models. We also currently apply the VE attention replacement across all heads in the selected middle layers. However, a more nuanced approach, which selectively replacing only those heads that are prone to hallucinations, may further optimize performance. Exploring the functionality of individual attention heads in these critical layers, and developing head-specific replacement strategies, represent promising directions for future work.
{
    \small
    \bibliographystyle{ieeenat_fullname}
    \bibliography{main}

\begin{thebibliography}{48}
\providecommand{\natexlab}[1]{#1}
\providecommand{\url}[1]{\texttt{#1}}
\expandafter\ifx\csname urlstyle\endcsname\relax
  \providecommand{\doi}[1]{doi: #1}\else
  \providecommand{\doi}{doi: \begingroup \urlstyle{rm}\Url}\fi

\bibitem[Achiam et~al.(2023)Achiam, Adler, Agarwal, Ahmad, Akkaya, Aleman, Almeida, Altenschmidt, Altman, Anadkat, et~al.]{achiam2023gpt}
Josh Achiam, Steven Adler, Sandhini Agarwal, Lama Ahmad, Ilge Akkaya, Florencia~Leoni Aleman, Diogo Almeida, Janko Altenschmidt, Sam Altman, Shyamal Anadkat, et~al.
\newblock Gpt-4 technical report.
\newblock \emph{arXiv preprint arXiv:2303.08774}, 2023.

\bibitem[Brown et~al.(2020)Brown, Mann, Ryder, Subbiah, Kaplan, Dhariwal, Neelakantan, Shyam, Sastry, Askell, et~al.]{brown2020language}
Tom Brown, Benjamin Mann, Nick Ryder, Melanie Subbiah, Jared~D Kaplan, Prafulla Dhariwal, Arvind Neelakantan, Pranav Shyam, Girish Sastry, Amanda Askell, et~al.
\newblock Language models are few-shot learners.
\newblock In \emph{Advances in neural information processing systems}, 2020.

\bibitem[Chen et~al.(2023)Chen, Zhang, Zeng, Zhang, Zhu, and Zhao]{chen2023shikra}
Keqin Chen, Zhao Zhang, Weili Zeng, Richong Zhang, Feng Zhu, and Rui Zhao.
\newblock Shikra: Unleashing multimodal llm's referential dialogue magic.
\newblock \emph{arXiv preprint arXiv:2306.15195}, 2023.

\bibitem[Chen et~al.(2024)Chen, Sikka, Cogswell, Ji, and Divakaran]{chen2024dress}
Yangyi Chen, Karan Sikka, Michael Cogswell, Heng Ji, and Ajay Divakaran.
\newblock Dress: Instructing large vision-language models to align and interact with humans via natural language feedback.
\newblock In \emph{Proceedings of the IEEE/CVF Conference on Computer Vision and Pattern Recognition}, pages 14239--14250, 2024.

\bibitem[Chiang et~al.(2023)Chiang, Li, Lin, Sheng, Wu, Zhang, Zheng, Zhuang, Zhuang, Gonzalez, et~al.]{chiang2023vicuna}
Wei-Lin Chiang, Zhuohan Li, Ziqing Lin, Ying Sheng, Zhanghao Wu, Hao Zhang, Lianmin Zheng, Siyuan Zhuang, Yonghao Zhuang, Joseph~E Gonzalez, et~al.
\newblock Vicuna: An open-source chatbot impressing gpt-4 with 90\%* chatgpt quality.
\newblock \emph{See https://vicuna. lmsys. org (accessed 14 April 2023)}, 2\penalty0 (3):\penalty0 6, 2023.

\bibitem[Cho et~al.(2025)Cho, Kim, Hwang, and Cho]{cho2025you}
Yeongjae Cho, Keonwoo Kim, Taebaek Hwang, and Sungzoon Cho.
\newblock Do you keep an eye on what i ask? mitigating multimodal hallucination via attention-guided ensemble decoding.
\newblock \emph{arXiv preprint arXiv:2505.17529}, 2025.

\bibitem[Cui et~al.(2024)Cui, Ma, Cao, Ye, Zhou, Liang, Chen, Lu, Yang, Liao, et~al.]{cui2024survey}
Can Cui, Yunsheng Ma, Xu Cao, Wenqian Ye, Yang Zhou, Kaizhao Liang, Jintai Chen, Juanwu Lu, Zichong Yang, Kuei-Da Liao, et~al.
\newblock A survey on multimodal large language models for autonomous driving.
\newblock In \emph{Proceedings of the IEEE/CVF winter conference on applications of computer vision}, pages 958--979, 2024.

\bibitem[Darcet et~al.(2023)Darcet, Oquab, Mairal, and Bojanowski]{darcet2023vision}
Timoth{\'e}e Darcet, Maxime Oquab, Julien Mairal, and Piotr Bojanowski.
\newblock Vision transformers need registers.
\newblock \emph{arXiv preprint arXiv:2309.16588}, 2023.

\bibitem[Fu et~al.(2025)Fu, Chen, Shen, Qin, Zhang, Lin, Yang, Zheng, Li, Sun, Wu, Ji, Shan, and He]{fu2025mme}
Chaoyou Fu, Peixian Chen, Yunhang Shen, Yulei Qin, Mengdan Zhang, Xu Lin, Jinrui Yang, Xiawu Zheng, Ke Li, Xing Sun, Yunsheng Wu, Rongrong Ji, Caifeng Shan, and Ran He.
\newblock {MME}: A comprehensive evaluation benchmark for multimodal large language models.
\newblock In \emph{The Thirty-ninth Annual Conference on Neural Information Processing Systems Datasets and Benchmarks Track}, 2025.

\bibitem[Fu et~al.(2024)Fu, Bonnen, Guillory, and Darrell]{fu2024hidden}
Stephanie Fu, Tyler Bonnen, Devin Guillory, and Trevor Darrell.
\newblock Hidden in plain sight: {VLMs} overlook their visual representations.
\newblock \emph{arXiv preprint arXiv:2406.05346}, 2024.

\bibitem[Gong et~al.(2024)Gong, Ming, Wang, and Wei]{gong2024damro}
Xuan Gong, Tianshi Ming, Xinpeng Wang, and Zhihua Wei.
\newblock Damro: Dive into the attention mechanism of lvlm to reduce object hallucination.
\newblock \emph{arXiv preprint arXiv:2410.04514}, 2024.

\bibitem[Gunjal et~al.(2024)Gunjal, Yin, and Bas]{gunjal2024detecting}
Anisha Gunjal, Jihan Yin, and Erhan Bas.
\newblock Detecting and preventing hallucinations in large vision language models.
\newblock In \emph{Proceedings of the AAAI Conference on Artificial Intelligence}, pages 18135--18143, 2024.

\bibitem[Hu et~al.(2023)Hu, Zhang, Zhao, and Sun]{hu2023ciem}
Hongyu Hu, Jiyuan Zhang, Minyi Zhao, and Zhenbang Sun.
\newblock Ciem: Contrastive instruction evaluation method for better instruction tuning.
\newblock \emph{arXiv preprint arXiv:2309.02301}, 2023.

\bibitem[Huang et~al.(2024)Huang, Dong, Zhang, Wang, He, Wang, Lin, Zhang, and Yu]{huang2024opera}
Qidong Huang, Xiaoyi Dong, Pan Zhang, Bin Wang, Conghui He, Jiaqi Wang, Dahua Lin, Weiming Zhang, and Nenghai Yu.
\newblock Opera: Alleviating hallucination in multi-modal large language models via over-trust penalty and retrospection-allocation.
\newblock In \emph{Proceedings of the IEEE/CVF Conference on Computer Vision and Pattern Recognition}, pages 13418--13427, 2024.

\bibitem[Jiang et~al.(2025)Jiang, Chen, Zhu, Luo, Shen, and Yang]{jiang2025devils}
Zhangqi Jiang, Junkai Chen, Beier Zhu, Tingjin Luo, Yankun Shen, and Xu Yang.
\newblock Devils in middle layers of large vision-language models: Interpreting, detecting and mitigating object hallucinations via attention lens.
\newblock In \emph{Proceedings of the Computer Vision and Pattern Recognition Conference}, pages 25004--25014, 2025.

\bibitem[Kim et~al.(2023)Kim, Koepke, Schmid, and Akata]{kim2023exposing}
Jae~Myung Kim, A Koepke, Cordelia Schmid, and Zeynep Akata.
\newblock Exposing and mitigating spurious correlations for cross-modal retrieval.
\newblock In \emph{Proceedings of the IEEE/CVF conference on computer vision and pattern recognition}, pages 2585--2595, 2023.

\bibitem[Leng et~al.(2024)Leng, Zhang, Chen, Li, Lu, Miao, and Bing]{leng2024mitigating}
Sicong Leng, Hang Zhang, Guanzheng Chen, Xin Li, Shijian Lu, Chunyan Miao, and Lidong Bing.
\newblock Mitigating object hallucinations in large vision-language models through visual contrastive decoding.
\newblock In \emph{Proceedings of the IEEE/CVF Conference on Computer Vision and Pattern Recognition}, pages 13872--13882, 2024.

\bibitem[Li et~al.(2022)Li, Li, Xiong, and Hoi]{li2022blip}
Junnan Li, Dongxu Li, Caiming Xiong, and Steven Hoi.
\newblock Blip: Bootstrapping language-image pre-training for unified vision-language understanding and generation.
\newblock In \emph{International Conference on Machine Learning}, 2022.

\bibitem[Li et~al.(2023{\natexlab{a}})Li, Li, Savarese, and Hoi]{li2023blip2}
Junnan Li, Dongxu Li, Silvio Savarese, and Steven Hoi.
\newblock Blip-2: Bootstrapping language-image pre-training with frozen image encoders and large language models.
\newblock \emph{arXiv preprint arXiv:2301.12597}, 2023{\natexlab{a}}.

\bibitem[Li et~al.(2019)Li, Yatskar, Yin, Hsieh, and Chang]{li2019visualbert}
Liunian~Harold Li, Mark Yatskar, Da Yin, Cho-Jui Hsieh, and Kai-Wei Chang.
\newblock Visualbert: A simple and performant baseline for vision and language.
\newblock In \emph{Proceedings of the 57th Annual Meeting of the Association for Computational Linguistics}, 2019.

\bibitem[Li et~al.(2023{\natexlab{b}})Li, Du, Zhou, Wang, Zhao, and Wen]{li2023evaluating}
Yifan Li, Yifan Du, Kun Zhou, Jinpeng Wang, Wayne~Xin Zhao, and Ji-Rong Wen.
\newblock Evaluating object hallucination in large vision-language models.
\newblock \emph{arXiv preprint arXiv:2305.10355}, 2023{\natexlab{b}}.

\bibitem[Li et~al.(2025)Li, Shi, Gao, Liu, Wang, Chen, Liu, Zhao, Wang, and Metaxas]{li2025hidden}
Zhuowei Li, Haizhou Shi, Yunhe Gao, Di Liu, Zhenting Wang, Yuxiao Chen, Ting Liu, Long Zhao, Hao Wang, and Dimitris~N Metaxas.
\newblock The hidden life of tokens: Reducing hallucination of large vision-language models via visual information steering.
\newblock \emph{arXiv preprint arXiv:2502.03628}, 2025.

\bibitem[Ling et~al.(2025)Ling, Wan, Jia, and Du]{ling2025driveblip2}
Shihong Ling, Yue Wan, Xiaowei Jia, and Na Du.
\newblock Driveblip2: Attention-guided explanation generation for complex driving scenarios.
\newblock \emph{arXiv preprint arXiv:2506.22494}, 2025.

\bibitem[Liu et~al.(2025)Liu, Kamath, Grunde-McLaughlin, Han, and Krishna]{liu2025visual}
Benlin Liu, Amita Kamath, Madeleine Grunde-McLaughlin, Winson Han, and Ranjay Krishna.
\newblock Visual representations inside the language model.
\newblock \emph{arXiv preprint arXiv:2510.04819}, 2025.

\bibitem[Liu et~al.(2023{\natexlab{a}})Liu, Lin, Li, Wang, Yacoob, and Wang]{liu2023mitigating}
Fuxiao Liu, Kevin Lin, Linjie Li, Jianfeng Wang, Yaser Yacoob, and Lijuan Wang.
\newblock Mitigating hallucination in large multi-modal models via robust instruction tuning.
\newblock \emph{arXiv preprint arXiv:2306.14565}, 2023{\natexlab{a}}.

\bibitem[Liu et~al.(2023{\natexlab{b}})Liu, Li, Wu, and Lee]{liu2023visual}
Haotian Liu, Chunyuan Li, Qingyang Wu, and Yong~Jae Lee.
\newblock Visual instruction tuning.
\newblock \emph{Advances in neural information processing systems}, 36:\penalty0 34892--34916, 2023{\natexlab{b}}.

\bibitem[Liu et~al.(2024{\natexlab{a}})Liu, Li, Li, and Lee]{liu2024improved}
Haotian Liu, Chunyuan Li, Yuheng Li, and Yong~Jae Lee.
\newblock Improved baselines with visual instruction tuning.
\newblock In \emph{Proceedings of the IEEE/CVF conference on computer vision and pattern recognition}, pages 26296--26306, 2024{\natexlab{a}}.

\bibitem[Liu et~al.(2024{\natexlab{b}})Liu, Xue, Chen, Chen, Zhao, Wang, Hou, Li, and Peng]{liu2024survey}
Hanchao Liu, Wenyuan Xue, Yifei Chen, Dapeng Chen, Xiutian Zhao, Ke Wang, Liping Hou, Rongjun Li, and Wei Peng.
\newblock A survey on hallucination in large vision-language models.
\newblock \emph{arXiv preprint arXiv:2402.00253}, 2024{\natexlab{b}}.

\bibitem[Liu et~al.(2024{\natexlab{c}})Liu, Wang, Cheng, and Li]{liu2024paying}
Shi Liu, Yifei Wang, Zhipeng Cheng, and Bo Li.
\newblock Paying more attention to image: A training-free method for alleviating hallucination in {LVLMs}.
\newblock In \emph{European Conference on Computer Vision}, 2024{\natexlab{c}}.

\bibitem[Liu et~al.(2024{\natexlab{d}})Liu, Ye, Xing, and Zou]{liu2024reducing}
Sheng Liu, Haotian Ye, Lei Xing, and James Zou.
\newblock Reducing hallucinations in vision-language models via latent space steering.
\newblock \emph{arXiv preprint arXiv:2410.15778}, 2024{\natexlab{d}}.

\bibitem[Rohrbach et~al.(2018)Rohrbach, Hendricks, Burns, Darrell, and Saenko]{rohrbach2018object}
Anna Rohrbach, Lisa~Anne Hendricks, Kaylee Burns, Trevor Darrell, and Kate Saenko.
\newblock Object hallucination in image captioning.
\newblock \emph{arXiv preprint arXiv:1809.02156}, 2018.

\bibitem[Sarkar et~al.(2025)Sarkar, Che, Gavin, Beerel, and Kundu]{sarkar2025mitigating}
Sreetama Sarkar, Yue Che, Alex Gavin, Peter~Anthony Beerel, and Souvik Kundu.
\newblock Mitigating hallucinations in vision-language models through image-guided head suppression.
\newblock In \emph{Proceedings of the 2025 Conference on Empirical Methods in Natural Language Processing}, pages 12492--12511, 2025.

\bibitem[Sun et~al.(2024)Sun, Shen, Cao, Liu, Li, Shen, Gan, Gui, Wang, Yang, et~al.]{sun2024aligning}
Zhiqing Sun, Sheng Shen, Shengcao Cao, Haotian Liu, Chunyuan Li, Yikang Shen, Chuang Gan, Liangyan Gui, Yu-Xiong Wang, Yiming Yang, et~al.
\newblock Aligning large multimodal models with factually augmented rlhf.
\newblock In \emph{Findings of the Association for Computational Linguistics: ACL 2024}, pages 13088--13110, 2024.

\bibitem[Touvron et~al.(2023)Touvron, Lavril, Izacard, Martinet, Lachaux, Lacroix, Rozi{\`e}re, Goyal, Hambro, Azhar, et~al.]{touvron2023llama}
Hugo Touvron, Thibaut Lavril, Gautier Izacard, Xavier Martinet, Marie-Anne Lachaux, Timoth{\'ee} Lacroix, Baptiste Rozi{\`e}re, Naman Goyal, Eric Hambro, Faisal Azhar, et~al.
\newblock Llama: Open and efficient foundation language models.
\newblock \emph{arXiv preprint arXiv:2302.13971}, 2023.

\bibitem[van~der Poel and Jijkoun(2022)]{van2022grit}
Jort van~der Poel and Vladimir Jijkoun.
\newblock {GRIT}: A generative region-to-text transformer for object grounding and question answering.
\newblock In \emph{Proceedings of the 2022 Conference of the North American Chapter of the Association for Computational Linguistics: Human Language Technologies}, 2022.

\bibitem[Vaswani et~al.(2017)Vaswani, Shazeer, Parmar, Uszkoreit, Jones, Gomez, Kaiser, and Polosukhin]{vaswani2017attention}
Ashish Vaswani, Noam Shazeer, Niki Parmar, Jakob Uszkoreit, Llion Jones, Aidan~N Gomez, {\L}ukasz Kaiser, and Illia Polosukhin.
\newblock Attention is all you need.
\newblock In \emph{Advances in neural information processing systems}, 2017.

\bibitem[Wan et~al.(2025)Wan, Zhang, Yong, Ma, Stepputtis, Morency, Ramanan, Sycara, and Xie]{wan2025only}
Zifu Wan, Ce Zhang, Silong Yong, Martin~Q Ma, Simon Stepputtis, Louis-Philippe Morency, Deva Ramanan, Katia Sycara, and Yaqi Xie.
\newblock Only: One-layer intervention sufficiently mitigates hallucinations in large vision-language models.
\newblock \emph{arXiv preprint arXiv:2507.00898}, 2025.

\bibitem[Wang et~al.(2024)Wang, Chen, Zhang, Tian, Xu, Deng, and Chen]{wang2024mllm}
Chenxi Wang, Xiang Chen, Ningyu Zhang, Bozhong Tian, Haoming Xu, Shumin Deng, and Huajun Chen.
\newblock Mllm can see? dynamic correction decoding for hallucination mitigation.
\newblock \emph{arXiv preprint arXiv:2410.11779}, 2024.

\bibitem[Wang et~al.(2023)Wang, Wang, Xu, Zhang, Gu, Jia, Wang, Xu, Yan, Zhang, et~al.]{wang2023amber}
Junyang Wang, Yuhang Wang, Guohai Xu, Jing Zhang, Yukai Gu, Haitao Jia, Jiaqi Wang, Haiyang Xu, Ming Yan, Ji Zhang, et~al.
\newblock Amber: An llm-free multi-dimensional benchmark for mllms hallucination evaluation.
\newblock \emph{arXiv preprint arXiv:2311.07397}, 2023.

\bibitem[Wang et~al.(2025)Wang, Gu, Gao, and Zhou]{wang2025damo}
Kaishen Wang, Hengrui Gu, Meijun Gao, and Kaixiong Zhou.
\newblock Damo: Decoding by accumulating activations momentum for mitigating hallucinations in vision-language models.
\newblock In \emph{The Thirteenth International Conference on Learning Representations}, 2025.

\bibitem[Wu et~al.(2023)Wu, Gan, Chen, Wan, and Yu]{wu2023multimodal}
Jiayang Wu, Wensheng Gan, Zefeng Chen, Shicheng Wan, and Philip~S Yu.
\newblock Multimodal large language models: A survey.
\newblock In \emph{2023 IEEE International Conference on Big Data (BigData)}, pages 2247--2256. IEEE, 2023.

\bibitem[Yao et~al.(2024)Yao, Li, Ren, Wang, Liu, Sun, and Hou]{yao2024deco}
Linli Yao, Lei Li, Shuhuai Ren, Lean Wang, Yuanxin Liu, Xu Sun, and Lu Hou.
\newblock Deco: Decoupling token compression from semantic abstraction in multimodal large language models.
\newblock \emph{arXiv preprint arXiv:2405.20985}, 2024.

\bibitem[Yin et~al.(2024{\natexlab{a}})Yin, Fu, Zhao, Li, Sun, Xu, and Chen]{yin2024survey}
Shukang Yin, Chaoyou Fu, Sirui Zhao, Ke Li, Xing Sun, Tong Xu, and Enhong Chen.
\newblock A survey on multimodal large language models.
\newblock \emph{National Science Review}, 11\penalty0 (12):\penalty0 nwae403, 2024{\natexlab{a}}.

\bibitem[Yin et~al.(2024{\natexlab{b}})Yin, Fu, Zhao, Xu, Wang, Sui, Shen, Li, Sun, and Chen]{yin2024woodpecker}
Shukang Yin, Chaoyou Fu, Sirui Zhao, Tong Xu, Hao Wang, Dianbo Sui, Yunhang Shen, Ke Li, Xing Sun, and Enhong Chen.
\newblock Woodpecker: Hallucination correction for multimodal large language models.
\newblock \emph{Science China Information Sciences}, 67\penalty0 (12):\penalty0 220105, 2024{\natexlab{b}}.

\bibitem[Yu et~al.(2024)Yu, Li, Wei, Pang, Ye, Qin, Tang, Tian, and Zhuang]{yu2024hallucidoctor}
Qifan Yu, Juncheng Li, Longhui Wei, Liang Pang, Wentao Ye, Bosheng Qin, Siliang Tang, Qi Tian, and Yueting Zhuang.
\newblock Hallucidoctor: Mitigating hallucinatory toxicity in visual instruction data.
\newblock In \emph{Proceedings of the IEEE/CVF Conference on Computer Vision and Pattern Recognition}, pages 12944--12953, 2024.

\bibitem[Zhang et~al.(2023)Zhang, Luo, Zhu, Li, Zeng, and Zhou]{zhang2023mitigating}
Yixuan Zhang, Yihan Luo, Yixuan Zhu, Wenhao Li, Qifan Zeng, and Ming Zhou.
\newblock Mitigating object hallucinations in large vision-language models through visual contrastive decoding.
\newblock In \emph{Proceedings of the 2023 Conference on Empirical Methods in Natural Language Processing}, 2023.

\bibitem[Zhou et~al.(2023)Zhou, Cui, Yoon, Zhang, Deng, Finn, Bansal, and Yao]{zhou2023analyzing}
Yiyang Zhou, Chenhang Cui, Jaehong Yoon, Linjun Zhang, Zhun Deng, Chelsea Finn, Mohit Bansal, and Huaxiu Yao.
\newblock Analyzing and mitigating object hallucination in large vision-language models.
\newblock \emph{arXiv preprint arXiv:2310.00754}, 2023.

\bibitem[Zhu et~al.(2023)Zhu, Chen, Shen, Li, and Elhoseiny]{zhu2023minigpt4}
Deyao Zhu, Jun Chen, Xiaoqian Shen, Xiang Li, and Mohamed Elhoseiny.
\newblock Minigpt-4: Enhancing vision-language understanding with advanced large language models.
\newblock \emph{arXiv preprint arXiv:2304.10592}, 2023.

\end{thebibliography}
}
\maketitlesupplementary

\begin{figure*}[ht]
  \centering
   \includegraphics[width=\linewidth]{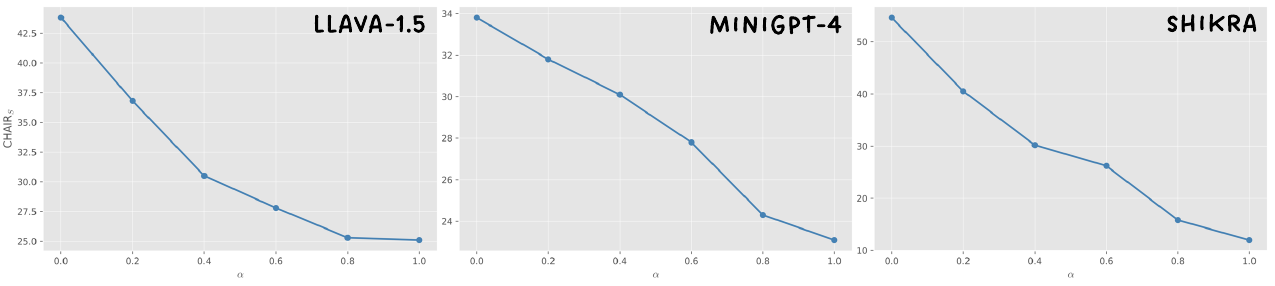}
   \caption{Ablation study on the logits weight $\alpha$ in using greedy decoding across three LVLM models on the CHAIR benchmark. For each model, we report performance when $\alpha$ is fixed to different values in the range $[0,1]$. The results illustrate how increasing $\alpha$, i.e., placing greater weight on the attention-replaced logits, impacts hallucination reduction.}
   \label{fig:ablation_alpha}
\end{figure*}

\section{Experiment Setups}
\label{sec:app_experiment_setup}
In VEGAS, we introduce adaptive logits steering based on the following vision attention block entropy (VABE):

\begin{equation}
\label{eq:vabe}
\mathrm{VABE}_4^l = \frac{1}{H} \sum_{h=1}^{H} \mathrm{BE}_4\!\left(\mat{\tilde{A}}^{(l,h)}_v\right),
\end{equation}
where $H$ is the number of attention heads and $\mat{\tilde{A}}^{(l,h)}_v$ denotes the pre-softmax attention over image tokens at head $h$ of layer $l$.  

The weighting coefficient $\alpha$ in the above logits steering is adaptively determined as:
\begin{equation}
\label{eq:adaptive_alpha}
    \alpha =
\begin{cases}
\alpha_{1}, & \text{if } \mathrm{VABE}_4^l > \eta,\\[4pt]
\alpha_{2}, & \text{otherwise},
\end{cases}
\end{equation}

In all experiments, we use $\mathrm{VABE}_4^{15}$ as the hallucination indicator. We choose $\eta=0.31$ for all LLAVA-1.5~\cite{liu2023visual} and Shikra~\cite{chen2023shikra}. For LLAVA-1.5 and MINIGPT-4~\cite{zhu2023minigpt4}, we use $\alpha_1=1.0$ and $\alpha_2=0.8$. And we set $\alpha_1=0.6$ and $\alpha_2=0.4$ for Shikra.

In MINIGPT-4, instead of using image tokens, query tokens are provided to the LLM as the vision inputs. Calculated from the self-attention and image token cross-attention, each query represents information from multiple image patches~\cite{yao2024deco,li2023blip2,ling2025driveblip2}. Thus for MINIGPT-4, we use query token attention entropy instead of VABE as the indicator. Specifically we choose $\eta=2.1$ as the threshold for query token attention entropy.

\section{Additional Experiments}

\subsection{Impact of the logits weight $\alpha$}
In this ablation study, we vary the weight parameter $\alpha$ to assess its impact on overall performance. We evaluate three LVLMs on the CHAIR benchmark using different fixed $\alpha\in[0,1]$. \cref{fig:ablation_alpha} presents the results. Generally, higher $\alpha$ values lead to better reductions in hallucinations. However, for Shikra, when $\alpha$ approaches 1, we observe extremely low object-hallucination rates but a tendency for the model to generate incomplete or truncated sentences. Accordingly, we adopt large $\alpha$ values for LLaVA-1.5 and MiniGPT-4, but a relatively smaller $\alpha$ for Shikra.

\begin{figure}[H]
  \centering
   \includegraphics[width=0.75\linewidth]{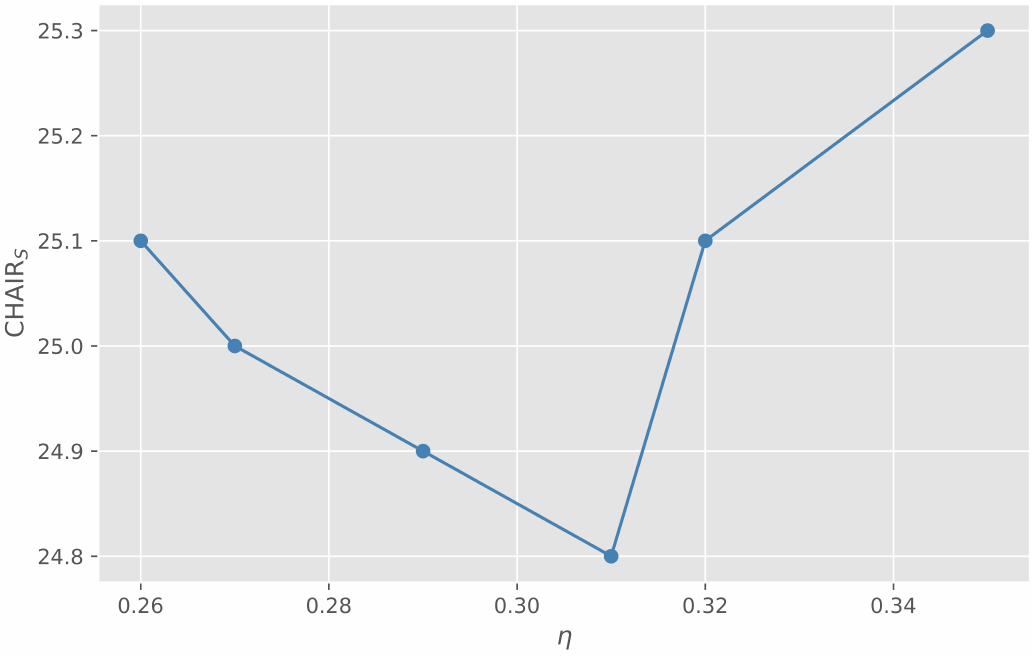}
   \caption{Ablation study on the threshold $\eta$ in VEGAS using greedy decoding on LLaVA-1.5-7B for the CHAIR benchmark. We vary $\eta$ across a range of values and observe how it affects performance: the optimal value is $\eta=0.31$. When $\eta$ is set too low or too high, VEGAS effectively behaves like a fixed $\alpha$ configuration ($\alpha=0.8$ or $\alpha=1.0$, respectively), which yields inferior results.}
   \label{fig:ablation_eta}
\end{figure}

\subsection{Ablation study on threshold $\eta$}
As described in \cref{sec:app_experiment_setup}, we apply different values of $\alpha$ depending on whether the current token’s $\mathrm{VABE}_4^{15}$ exceeds a threshold $\eta$. This technique, which we term Adaptive Logits Steering, enables dynamic weighting of the original and attention-replaced logits. In our ablation study, we vary $\eta$ and evaluate performance on the CHAIR task. \cref{fig:ablation_eta} shows that the optimal value is $\eta=0.31$. When $\eta$ is set much lower, the method defaults to $\alpha=0.8$; when $\eta$ is too high, it effectively behaves like fixed $\alpha=1.0$, in both cases yielding inferior performance.

\begin{figure}[H]
  \centering
   \includegraphics[width=0.9\linewidth]{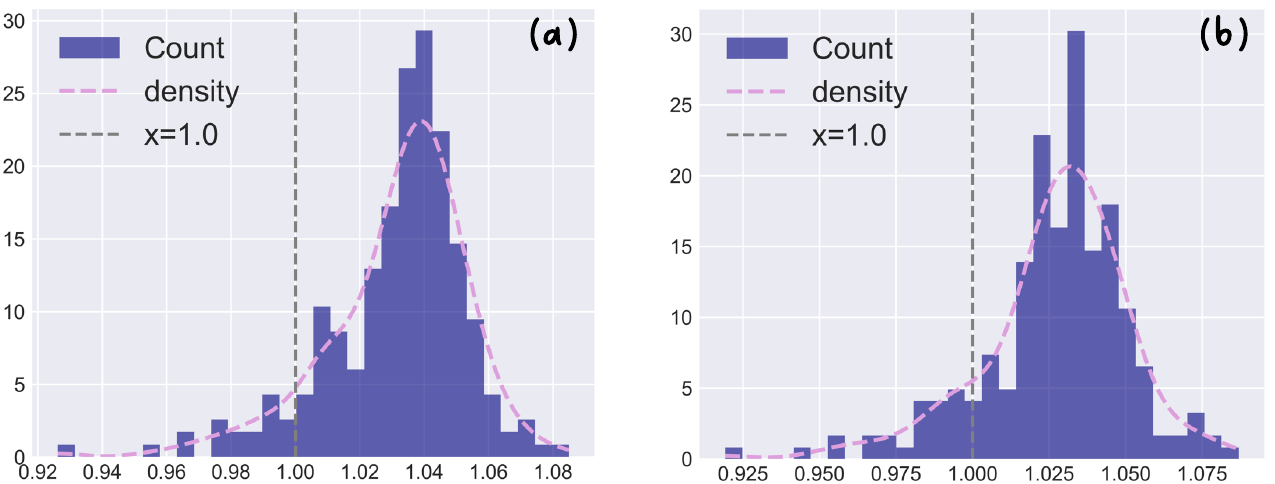}
   \caption{Ratio of hallucinated-token to non-hallucinated-token block entropy (\(\mathrm{BE}_4\)) for vision-attention maps in LLAVA-1.5. (a) Layer 15, (b) Layer 31. All values are calculated on real-object tokens. Ratios above 1.0 indicate that hallucinated tokens tend to exhibit higher block entropy. This pattern holds consistently across many layers within the LLM, including middle layers and final layers.}
   \label{fig:indicator_layers}
\end{figure}

\subsection{Hallucination indicator at various layers}
As described in \cref{sec:app_experiment_setup}, we use $\mathrm{VABE}_4^{15}$ as our primary hallucination indicator. \cref{fig:indicator_layers} shows that many layers within the LLM (including both middle and final layers) can serve as effective indicators of hallucination risk. To maintain simplicity and consistency in our framework, we thus adopt $\mathrm{VABE}_4^{15}$ as the default indicator throughout.

\begin{figure}[H]
  \centering
   \includegraphics[width=0.9\linewidth]{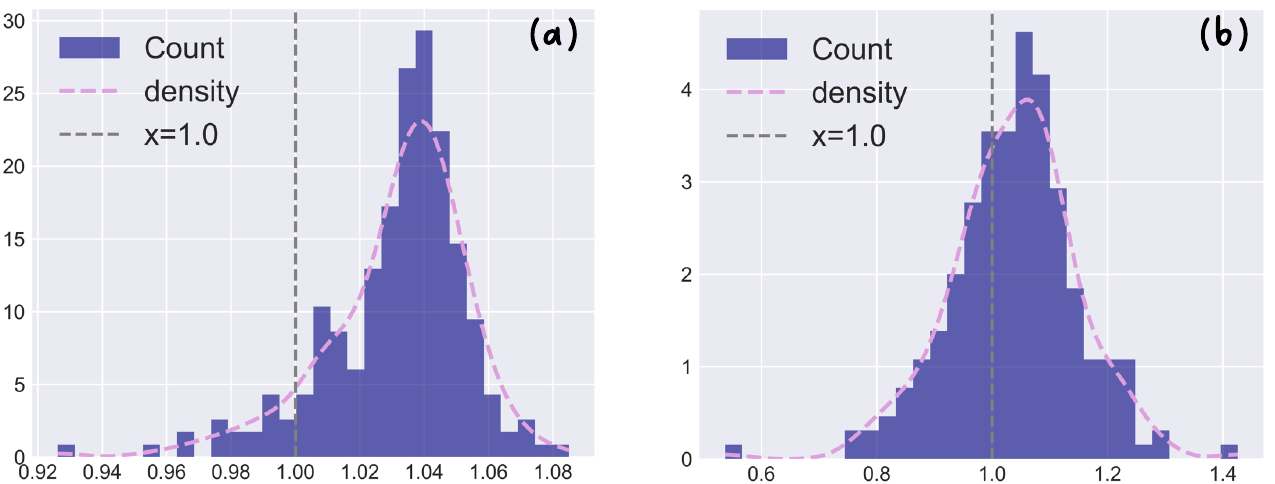}
   \caption{Ratio of hallucinated-token to non-hallucinated-token vision-attention block entropy (\(\mathrm{VABE}\)) in LLAVA-1.5 at Layer 15: (a) calculated using \(\mathrm{VABE}_4^{15}\); (b) calculated using \(\mathrm{VABE}_8^{15}\). All values are calculated on real-object tokens. The ratio typically exceeds 1.0 when using \(\mathrm{VABE}_4^{15}\), indicating that hallucinated tokens tend to exhibit larger \(\mathrm{VABE}_4^{15}\). However, when using a larger block size (e.g., 8), \(\mathrm{VABE}\) no longer clearly differentiates between hallucinated and non-hallucinated tokens.
}
   \label{fig:indicator_block_size}
\end{figure}

\subsection{VABE block size in hallucination detection}
As defined in \cref{eq:vabe}, \(\mathrm{VABE}\) is derived from our introduced Block Entropy metric, where the choice of block size is a critical hyperparameter. Accordingly, we evaluate the effect of varying block size on the effectiveness of the hallucination indicator. As shown in \cref{fig:indicator_block_size}, smaller block sizes (e.g., 4) provide clearer discrimination between hallucinated and non-hallucinated tokens.


\end{document}